\documentclass[review]{elsarticle}
\usepackage{amssymb}
\usepackage{latexsym}
\usepackage{subfigure}
\usepackage{graphicx}
\usepackage{diagbox}
\usepackage{url}
\usepackage{xcolor}
\usepackage{amsmath}
\usepackage{framed,multirow}
\usepackage{bm}
\usepackage{lineno,hyperref}
\usepackage{algorithm}
\usepackage{algorithmic}
\usepackage{setspace}
\modulolinenumbers[5]

\journal{Journal of \LaTeX\ Templates}

\begin{document}

\begin{frontmatter}

\title{A Feature Consistency Driven Attention Erasing Network for Fine-Grained Image Retrieval}

%% Group authors per affiliation:
% \author{Elsevier\fnref{myfootnote}}
% \address{Radarweg 29, Amsterdam}
% \fntext[myfootnote]{Since 1880.}

%% or include affiliations in footnotes:
\author[mymainaddress]{Qi Zhao}
\ead[url]{zhaoqi@buaa.edu.cn}
\author[mymainaddress]{Xu Wang}
\ead[url]{sy2002406@buaa.edu.cn}
\author[mymainaddress]{Shuchang Lyu}
\ead[url]{lyushuchang@buaa.edu.cn}
\author[mymainaddress]{Binghao Liu}
\ead[url]{liubinghao@buaa.edu.cn}

% corresponding author
\author[mymainaddress]{Yifan Yang\corref{mycorrespondingauthor}}
\cortext[mycorrespondingauthor]{Corresponding author}
\ead{stephenyoung@buaa.edu.cn}

\address[mymainaddress]{Beihang University, Beijing, China}

\begin{abstract}
\par
Large-scale fine-grained image retrieval has two main problems. First, low dimensional feature embedding can fasten the retrieval process but bring accuracy reduce due to overlooking the feature of significant attention regions of images in fine-grained datasets. Second, fine-grained images lead to the same category query hash codes mapping into the different cluster in database hash latent space. To handle these two issues, we propose a feature consistency driven attention erasing network (FCAENet) for fine-grained image retrieval. For the first issue, we propose an adaptive augmentation module in FCAENet, which is selective region erasing module (SREM). SREM makes the network more robust on subtle differences of fine-grained task by adaptively covering some regions of raw images. The feature extractor and hash layer can learn more representative hash code for fine-grained images by SREM. With regard to the second issue, we fully exploit the pair-wise similarity information and add the enhancing space relation loss (ESRL) in FCAENet to make the vulnerable relation stabler between the query hash code and database hash code. We conduct extensive experiments on five fine-grained benchmark datasets (CUB2011, Aircraft, NABirds, VegFru, Food101) for 12bits, 24bits, 32bits, 48bits hash code. The results show that FCAENet achieves the state-of-the-art (SOTA) fine-grained retrieval performance compared with other methods.     

\end{abstract}


\end{frontmatter}

\section{Introduction}
%\paragraph{Installation}

Deep hash learning based image retrieval methods \cite{cui2020exchnet, jin2020deep} have achieved a huge success recent year. \cite{cui2020exchnet} solves on the fine-grained hashing problem by a feature exchanging method. It aims to learn compact hash code for specific fine-grained object via an end-to-end architecture. Besides, \cite{jin2020deep} find salient regions and learn to preserve semantic information for addressing the limitation of hashing code representing the fine-grained object. Although the previous hash learning based methods improve the representative ability of low dimensional hash code for fine-grained image retrieval, how to obtain more representative hash code for fine-grained images still need to be explored.  

\par
\begin{figure}[]
    \centering
    \includegraphics[width=1.0\linewidth]{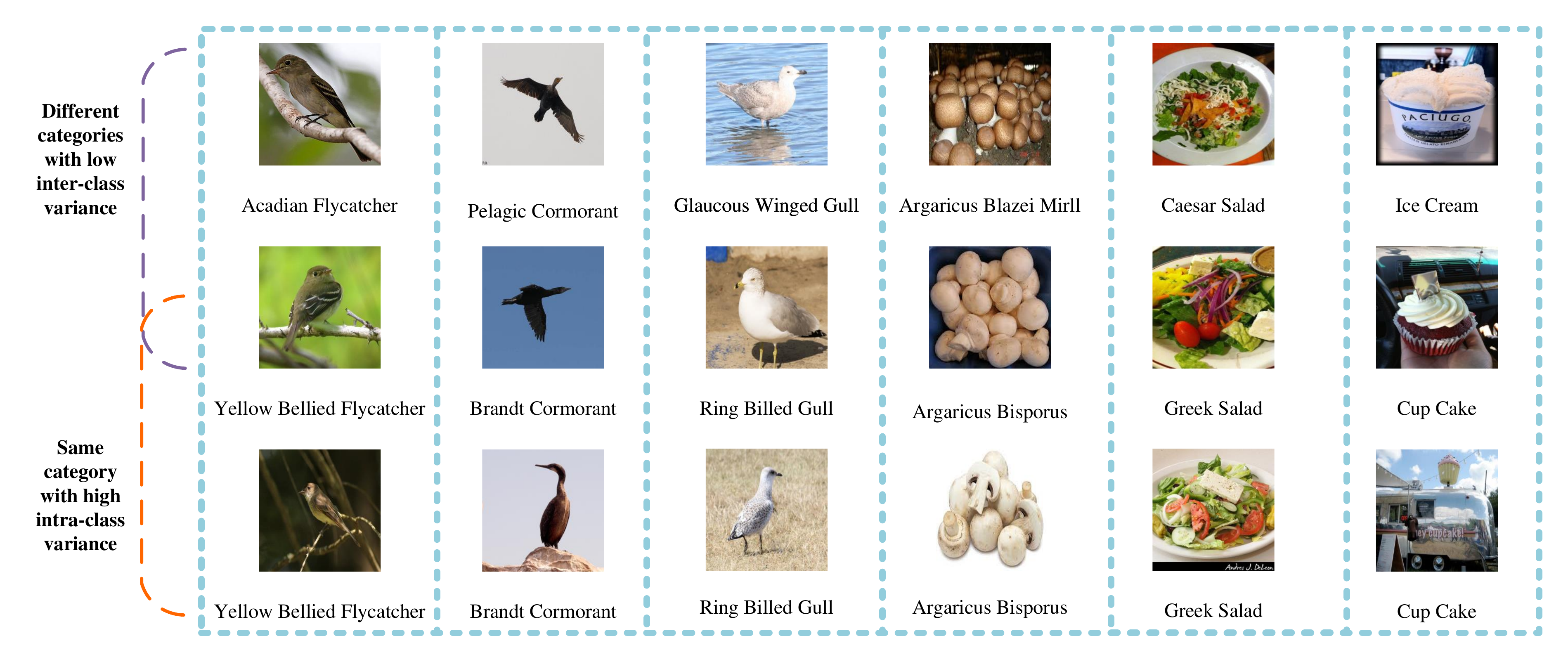}
    \caption{Intuitive description about main challenges of fine-grained task. There are six columns in this figure. Each column represents a set of the high intra-class variance and low inter-class variance images. These challenges seriously decrease the performance of retrieval accuracy.}
    \label{Fig.1}
\end{figure}

As for the fine-grained image retrieval, there are two basic components: query hash code, database hash code. To obtain these two hash codes, many efforts \cite{liu2016deep, jiang2018asymmetric, cui2020exchnet} have been done via symmetric hash learning methods or asymmetric hash learning methods. Deep Supervised Hashing \cite{liu2016deep} applies the negative-pair and positive-pair to encourage the output hash code to approximate discrete values. \cite{jiang2018asymmetric} proposes an asymmetric deep hashing method and obviously improves the training efficiency and retrieval accuracy by asymmetrically learning pair-wise similarity. \cite{cui2020exchnet} researches on the fine-grained hashing problem, which aims to learn compact hash code for specific fine-grained object via an end-to-end architecture. By contrast to the symmetric hash learning methods, asymmetric hash learning methods show the outstanding retrieval performance and training efficiency. However, there are two main problems in fine-grained image retrieval. Simply representing the significant regions of fine-grained images (Fig.~\ref{Fig.1}) by low dimensional hash code is difficult, because the low dimensional hash code reduces much essential information of raw fine-grained images. Learning by an asymmetric method for query hash code and database hash code is vulnerable, because of the fine-grained images common properties that are the low inter-class variance and high intra-class variance. This properties may result the same category images mapping into the different clusters in hash latent space.

\par
\begin{figure}
    \centering
    \includegraphics[width=12cm, height=6.5cm]{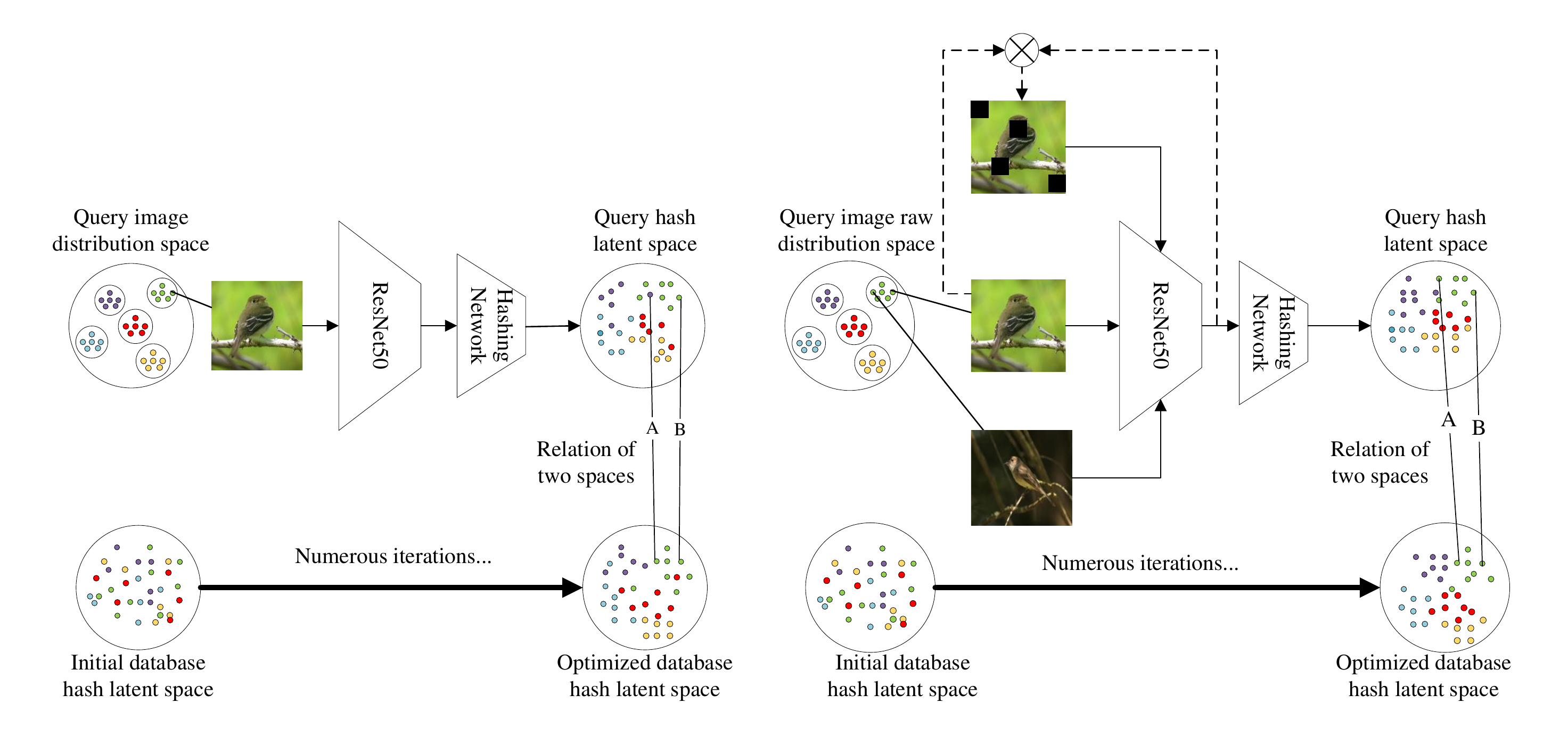}
    \caption{Comparison between the previous asymmetric hashing learning paradigm and ours. The left part is ADSH paradigm and the right part is FCAENet paradigm. Based on the pair-wise information and adaptively erasing data augmentation method, we construct triplets to integrate local and global feature better in hash latent space. In the other hand, we enhance the relation between query hash latent space and database latent space by making the positive pairs can mapping to the same cluster in hash latent space.}
    \label{Fig.2}
\end{figure}

 Motivated by solving the above mentioned two main problems, we propose a novel end-to-end trainable architecture named FCAENet. Different from the previous \cite{jiang2018asymmetric} paradigm, FCAENet paradigm is shown in Fig.~\ref{Fig.2}. Specifically, FCAENet consists of four components: feature extractor, selective region erasing module (SREM), hash layer, enhance space relation loss (ESRL). Feature extractor is applied to represent the discriminative features of a query image. We employ the SREM to make the network extract more representative hash code for raw images, so that we can improve the poor representation ability of low dimensional hash code. Additionally, for solving the problem that the same category images map to different clusters in hash code space, we introduce the ESRL to strengthen the relation between query hash code space and database hash code space based on the ADSH \cite{jiang2018asymmetric} loss function. Finally, we use hash layer to generate the hash code. By the above methods, we achieve the promising retrieval accuracy for fine-grained images.
\par The contributions of our paper can be summarized as follows:
\begin{itemize}
\item We design an end-to-end architecture (FCAENet) to learn a more representative hash code and preserve pair-wise similarity better than previous hashing based fine-grained image retrieval methods.
\item We propose a novel data augmentation method named selective region erasing module (SREM) to make the feature extractor more robust for large-scale fine-grained image retrieval. 
\item We introduce the enhance space relation loss (ESRL) as a penalty item to make the query hash code more relative to the database hash code.
\item We conduct abundant experiments on fine-grained datasets and achieve the state-of-the-art (SOTA) results for fine-grained image retrieval.
\end{itemize}

The rest of this paper is organized as follows. Sec.~\ref{section2} summarizes the recent related works. Sec.~\ref{section3} introduces our methods in detail. Sec.~\ref{section4} shows our experimental results and analyzes the function of our methods. Sec.~\ref{section5} is the conclusion part of our paper.

\section{Related work}
\label{section2}

{\bf Fine-Grained Image Recognition.} Fine-Grained task is a popular topic in many fields, such as image classification, object detection, image retrieval, instance segmentation and so on. Fine-grained image recognition focuses on subtle parts that can discriminate the similar categories.
\par Fine-grained image recognition method can divide into two categories. The first category \cite{zhang2014part,branson2014improved} is based on the parts annotation to drive the model learning more representative features. It boosts the previous methods \cite{duan2012discovering,yao2012codebook} on fine-grained classification. \cite{zhang2014part} introduces the CNN-Based Detector to obtain some semantic part information for object recognition. \cite{branson2014improved} introduces a compact pose normalization space to localize image patches features for fine-grained classification.  
\par  However, all the part-based fine-grained methods have a serious defect that they acquire part-level annotations. To tackle above limitation, \cite{krause2015fine} exploits co-segmentation and alignment methods to generate discriminative regions aiming to learn which parts are vital to recognition. \cite{huang2019attention,yang2018learning,gao2020channel,behera2021context} researches on attention-based methods that just exploits image-level annotations to guide the deep network focusing on the most essential region with regard to each category. \cite{yang2018learning} designs a training paradigm that enables navigator to detect most informative regions via the teacher. \cite{gao2020channel} explores the rich relationships between self-channels and interaction-channels which assists the network obtaining subtle inter-class difference. \cite{behera2021context} adopts a context-aware attention pooling to capture subtle changes via sub-pixel gradients.
\par
{\bf Fine-Grained Image Retrieval.}
Image retrieval is always a significant filed in computer vision. Traditional image retrieval methods such as \cite{jegou2011aggregating} are based on integration local feature strategy. CNN has shown its outstanding performance for recognizing the patterns. With few exceptions, image retrieval is closely linked with deep learning. Considerable researches (e.g.\cite{deng2018triplet, wang2020deep, zhao2015deep, cao2016quartet, gordo2017beyond, wei2017selective, jiang2018asymmetric}) on deep learning based image retrieval can be categorized to cross-modal \cite{deng2018triplet}, sketch-based \cite{wang2020deep}, Multi-label \cite{zhao2015deep}, Instance \cite{cao2016quartet}, Object \cite{chen2014ranking}, Semantic \cite{gordo2017beyond}, Fine-grained \cite{wei2017selective} and Asymmetric \cite{jiang2018asymmetric} according to the image retrieval type. Although CNN-based method has achieved an incredible success compared to classical image matching algorithm, confusing object pattern and complex background information are burning issues on image retrieval.
\par With the expansion of the retrieval image database, image retrieval task encounters the problem which is difficult to get correct retrieval results from fine-grained datasets. FGIR exactly focuses on eliminating this general fine-grained problem. \cite{xie2015fine} proposes the fine-grained image retrieval problem and explicit the hierarchy image retrieval database. Based on \cite{xie2015fine}, SCDA \cite{wei2017selective} exploits a pre-trained model to localize the major object region and irrelevant noisy parts. Then flood-fill algorithm is used to filter the noise and made the feature embeddings more representative. Aiming to build a generic local deep representation, \cite{zhou2016learning} more intensively analyses the role of global average pooling in CNNs and introduces the class activation mapping to highlight the discriminative regions. Based on \cite{selvaraju2017grad}, \cite{li2019guided} devises two streams end-to-end net structure to train a more convincing feature descriptor so that it directly guides the network feeding back for various tasks. \cite{cui2020exchnet} proposes local feature alignment and anchor-based learning pattern to generate discriminative representation. Finally, taking account of the query speed and calculation consumption, hashing learning method has become an indispensable part of real-time image retrieval, especially ADSH \cite{jiang2018asymmetric}.

\par Hash learning has been studied for approximate nearest neighbor search via mapping the high dimensional feature embedding to low dimensional hash code. Locality sensitive hashing(LSH) \cite{datar2004locality} and learning to hash \cite{weiss2008spectral, jegou2011aggregating} are two main methods in the light of data-dependent or not \cite{wang2017survey}. Ample experiments prove that LSH methods are inferior to learning to hash based methods, especially which integrates the deep learning method. Current hashing learning method based CNNs \cite{xia2014supervised, li2015feature, liu2016deep, zhu2016deep, cao2017hashnet, jiang2018asymmetric, cui2020exchnet, jin2020deep} mainly can be divided into supervised hashing learning and unsupervised hash learning respectively. \cite{xia2014supervised} employs feature extractor network to gain feature vectors and then uses encoder network to obtain hash code. \cite{li2015feature, liu2016deep, cao2017hashnet, jiang2018asymmetric} devises the end-to-end network encoding the images to low dimensional hash code. Different from the symmetric hash method \cite{li2015feature, liu2016deep, zhu2016deep}, \cite{jiang2018asymmetric} proposed an asymmetric method to accelerate training a model and achieve more promising performance than most symmetric hashing methods. \cite{jin2020deep} first combines the deep hashing with salient detection method to mine salient regions, which contains much semantic information. Although deep hashing method incredibly boots the image retrieval algorithm to application, semantic information grievously reduces which leads to the poor retrieval accuracy, especially for fine-grained image retrieval.

{\bf Representation learning.} Representation learning methods mainly revolve round devising pretext task, dividing into hand-designed pretext task method \cite{doersch2015unsupervised, chen2021jigsaw} and contrastive learning-based pretext task \cite{chen2020simple, chen2021exploring, he2020momentum, ye2020augmentation, henaff2020data, caron2020unsupervised}.
\par Jigsaw puzzles is an exemplary hand-designed pretext task so that considerable representation learning studies are proposed to analyse this topic. \cite{chen2021jigsaw} integrates the advantages of solving jigsaw puzzles and contrastive learning so that it accurately guides feature extractor via intra-image and inter-image information. Unlike the hand-designed pretext task, \cite{ye2020augmentation} exploits an data augmentation based unsupervised embedding method which aims to gain more discriminative feature representations in latent space. Specifically, pull the visually similar samples closer and push away the dissimilar samples in the embedding space. \cite{chen2020simple} proposes a simple yet effective contrastive self-supervised learning algorithm which devises multiple data augmentation operations and introduces an adaptive nonlinear transformation to promote the robustness of unsupervised learning feature extractor. \cite{chen2021exploring} discards the negative sample pairs, large batches and momentum encoder in previous Siamese architecture and discovers that stop-gradient training strategy can substantially prevent the simple network collapsing. The contrastive-based representation learning are inseparable from the essential motivation that minimize the gap between multiple augmentation of the same data and guide the network learning an appropriate parameter space.

\section{Method}
\label{section3}
In this section, we propose the FCAENet. Firstly, we illustrate what the hashing based fine-grained image retrieval is. Then, we present the SREM and ESRL. Learning to hash method is explained finally.

\begin{figure}
    \centering
    \includegraphics[width=12cm, height=6cm]{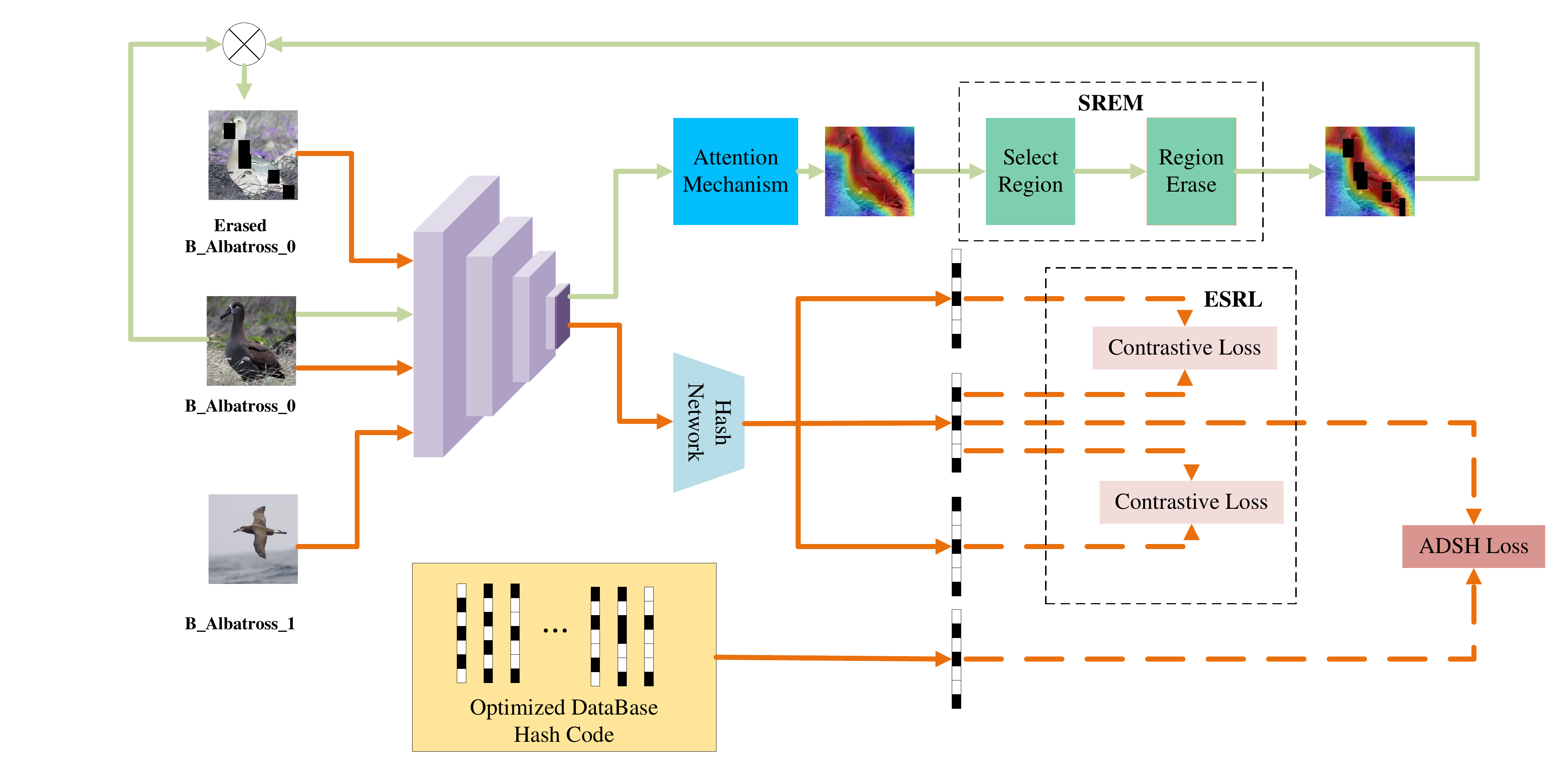}
    \caption{Structure of FCAENet. FCAENet contains two streams: light green stream is the adaptive augmentation branch to generate the erasing images; orange stream is the main fine-grained hashing process. In adaptive augmentation stream, the high-level features of CNN backbone are input to attention mechanism and obtain preliminary attention mask. Through the SREM, we adaptively erase some highlight regions and generate the erased images. Based on the previous stream and pair-wise similarity information, we construct the triplets as the input of main stream. The triplets successively go through CNN backbone and hash network to get compact hash code for fine-grained hashing.}
    \label{Fig.3}
\end{figure}

\subsection{Problem definition}
Image retrieval is proposed to search the same category images from a large scale database via a query image. Assume that we have ${m}$ query images which are denoted as ${X},{X} = \left \{{{x_i}} \right\}_{i=1}^{m} $ and ${n}$ database images which are denoted as ${Y}, {Y} = \left\{{{y_j}} \right\}_{j=1}^{n}$. Then, the supervised information is pair-wise similarity matrix which is denoted as ${ S} \in \left \{-1,+1 \right\}^{{m} \times {n}}$. If query images ${x_i}$ and database images ${y_j}$ are the same category, ${s_i{}_j}=1$, otherwise ${s_i{}_j}=-1$. Our goal is that makes the two hash codes preserve the raw pair-wise similarity as much as possible. Furthermore, we obtain the $k$bits query hash code  $ U = \left \{  {u_i} \right\}_{i=1}^{m}$ and $k$bits database hash code  $ V = \left \{  {v_j} \right\}_{j=1}^{n}$ by network or optimization algorithm. In order to metric the pair-wise similarity, we exploit Hamming distance \emph{d($ {u_i}$, $ {v_j}$)}. The fine-grained datasets have been split into train dataset and test dataset. So we appoint the test dataset as the query images for inference, all of the train dataset as the database images and sample $r$ images from train dataset for training process.

\subsection{The architecture of FCAENet}
As is shown in Fig.~\ref{Fig.3}, FCAENet includes four components. The first component is the feature extractor based on CNN. During the training process, we are given an image ${x_i}$ from the train-set. By the pair-wise similarity information, we select the positive-pair for $ {x_i}$ as ${x_i{}^p}$. Employing the attention mechanism and SREM, we obtain the data augmentation of the ${x_i}$ which denote as ${\tilde{x_i}}$ . Through the above operation, we build a triplet  $ x_{tri}=\left \{{x_i,x_i{}^p,\tilde{x_i}}\right\}$ as the input of FCAENet. However, during the inference process, we just input ${x_i}$ for promoting the retrieval speed in large-scale database. $ x_{tri}$ flows to the shared backbone and output three feature embeddings in Eq.~\ref{eq1}.
\begin{equation}
    {z_{tri}} = f({x_{tri}}),
    \label{eq1}
\end{equation}
Where $ z_{tri}=\left\{{{z_i},{z_i{}^p},{\tilde{z_i}}} \right\}$ denotes the feature embeddings of the $ x_{tri}$. And \emph {$f(\cdot)$} denotes the CNN backbone.
\par The second component is the hash layer. $z_{tri}$ as the input of this part is first mapped to the low dimensional triplet vectors and then they are activated by non-linear function to generate hash codes for image retrieval in Eq.~\ref{eq2}.
\begin{equation}
    u_{tri} =h(z_{tri})=\mathbf{\sigma}(W(z_{tri})),
    \label{eq2}
\end{equation}
Where $ u_{tri}=\left\{{{u_i},{u_i{}^p},{\tilde{u_i}}} \right\}$ denotes the final output hash code. \emph{$ h(\cdot)$} denotes the hash layer. \emph{$\sigma(\cdot)$} is the activation function such as \emph{$\tanh$}. \emph{W($\cdot$)} denotes a linear transformation to mapping the high dimension feature vector to low dimension.
The third component is learning compact hash code for database. After the $u_{tri}$ obtained from the hash layer, we optimize the randomly generating hash code of database bit by bit with the fixed $ u_{i}$. With this training strategy, we effectively achieve the aim of preserving the similarity between query hash code and database hash code.

\par During the CNN backbone extracting $ z_{i}$ from $ x_{i}$, we take the last convolution layer output as our concerned feature map denoted as $ A_{i}$. The last component of FCAENet consists of attention mechanism and SREM. $ A_{i}$ is the input of the attention mechanism to get the class activation map for SREM. The final output of SREM is an aligned mask which will be used for producing the $ {\tilde{x_{i}}}$ via making element-wise product between the mask and $ x_{i}$. The details of attention mechanism and SREM particularly introduce in next part.     

\begin{figure}
    \centering
    \includegraphics[]{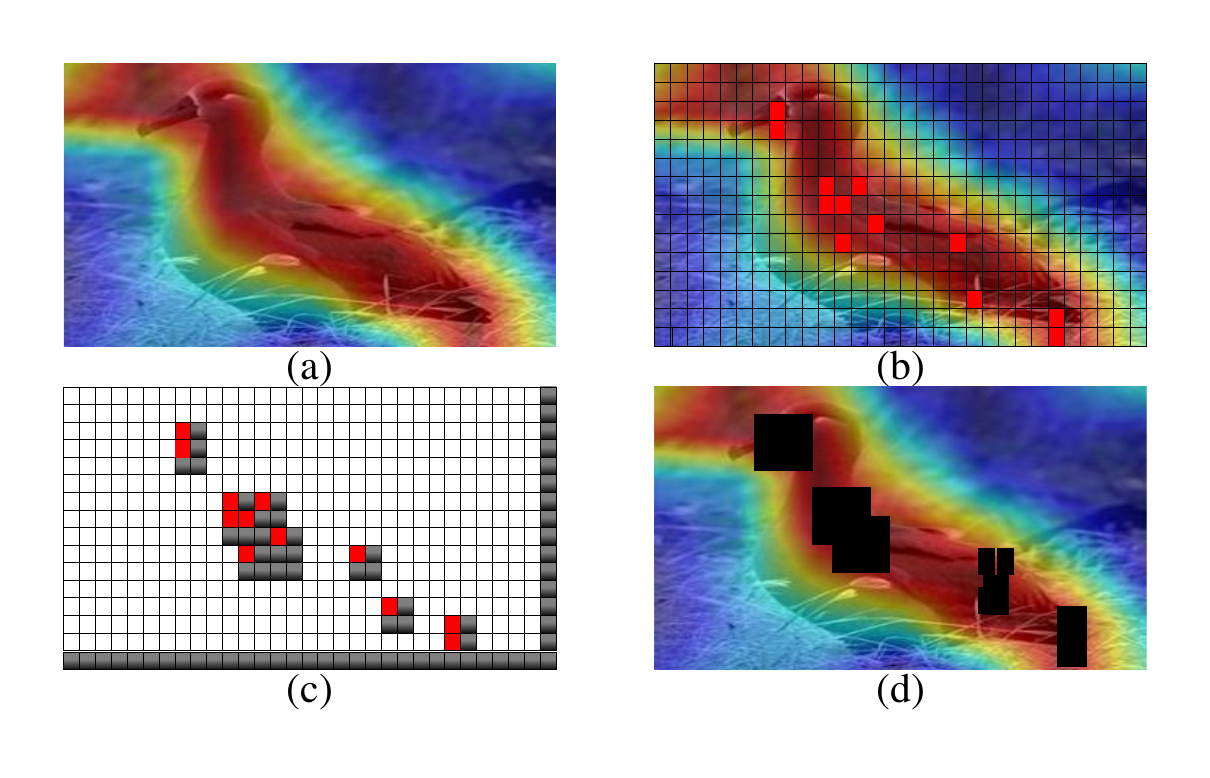}
    \caption{The illustration of the SREM. The Fig(a) shows the output of the attention mechanism. After we obtain the preliminary attention mask, we select the position of top n values which is shown as the red dots of Fig(b). Then we appoint position of red dots as the left top position of erasing region as is shown in Fig(c). The Fig(d) is the output of the SREM.}
    \label{Fig.4}
\end{figure}

\subsection{Selective region erasing module}
The performance in the fine-grained visual tasks always degenerate severely than coarse, as a consequence of which the subtly discriminative details and complex background carry serious noise to representation learning network.To alleviate this problem, we propose SREM which is shown in Fig.~\ref{Fig.4}.

\par $A_i$ is the input of the attention mechanism. Note that $ x_{i} \in {\Re ^{c\times h \times w}}$ and $A_{i} \in {\Re ^{c^\prime \times h^\prime \times w^\prime}}$. Each channel represents the different significance for the specific image cluster. Hence, we calculate the mean values $ {\bar{A_i}} \in {\Re ^{h^\prime \times w^\prime}}$ in channel dimension in Eq.~\ref{eq3}.
\begin{equation}
    {\bar{A_i}} = \frac{\sum_{k=1}^{c^\prime} {A_{ik}}}{c^\prime}
    \label{eq3}
\end{equation}
Where $ {A_{ik}}$ denotes the $k$-th channel of the $ {A_{i}{}}$. Then the bilinear interpolation is employed to reshaping $ \bar{A_i}$ to $ {\tilde{A_i}} \in {\Re ^{h \times w}}$. Additionally, in order to achieve the data augmentation operation on raw image, we normalize the $ {\tilde{A_i}}$ to $ {{M_i}} \in {\Re ^{h \times w}}$ whose values are limited between the $(\epsilon ,1+\epsilon)$ in Eq.~\ref{eq4}.
\begin{equation}
    {M_i} = \frac{\tilde{A_i}-min(\tilde{A_i})}{max(\tilde{A_i})-min(\tilde{A_i})} + {\epsilon}
    \label{eq4}
\end{equation}
The $\epsilon$ denotes a minimal nonzero. SREM is shown in Fig.~\ref{Fig.4}. Based on the coarse attention mask $ {\tilde{M_i}}$, we pad ${\tilde{M_i}}$ to $ {\bar{M_i}}\in {\Re ^{(h+l) \times (w+l)}}$ for the right and the bottom side with zero. Then, rank the $ {\bar{M_i}}$ as a descending order and confirm where ${top n}$ values are so that we start to erase the ${l \times l}$ grids from the left top pixels based on the positions of the $ {top n}$ values. Specifically, the algorithm of SREM is in Algorithm.\ref{alg1}.

\renewcommand{\algorithmicrequire}{\textbf{Input:}}
\renewcommand{\algorithmicensure}{\textbf{Output:}}

\begin{algorithm}[t]
%\setstretch{1.35}
\caption{selective region erasing algorithm}
\label{alg1}
\begin{algorithmic}
\REQUIRE Raw attention mask from attention mechanism, ${{M_i}}$; the size of the erasers, ${\emph{l}}$. 
the number of the erasers, ${\emph{n}}$;
\ENSURE Aligned attention mask, ${\tilde{M_i}} \in {\left\{0,1 \right\} ^{h \times w}}$.
\STATE Rank ${{M_i}}$ as descending order;
\STATE Choose top n pixel position $P=\left\{{(x_s,y_s)}\right\}, s=1...n$;
\STATE Pad ${{M_i}} \in {\Re^{h\times w}}$ to ${\bar {M_i}} \in {\Re^{(h+l) \times (w+l)}}$;
\STATE Binarization ${\bar{M_i}}$:
\FOR{${{x}}$ = $1,...w+l$}
    \FOR{${{y}}$ = $1,...h+l$}
        \IF{${(x,y)}\in P $}
            \STATE ${\bar{M_i}{(x,y)}}=0$
        \ELSE
            \STATE ${\bar{M_i}{(x,y)}}=1$
        \ENDIF
    \ENDFOR
\ENDFOR
\STATE Erase selected region:
\FOR{${(x,y)}\in P $}
    \FOR{${{j}}$ = $1,...l$}
        \FOR{${{k}}$ = $1,...l$}
            \STATE ${\bar{M_i}{(x+j, y+k)}}=0$
        \ENDFOR
    \ENDFOR
\ENDFOR
\STATE Take out aligned mask: ${\tilde{M_i}}$ = ${\bar{M_i}}[0:w, 0:h]$
\end{algorithmic}
\end{algorithm}

After obtaining the aligned mask ${\tilde{M_i}}$ based on the
SREM, we combine input image $ x_{i}$ with ${{\tilde{M_i}}}$ in Eq.~\ref{eq5} to generate ${\tilde{x_i}}$ as the adaptive data augmentation result for constructing triplet.
\begin{equation}
    {\tilde{x_i}} = {\tilde{M_i}} \otimes {x_i}
    \label{eq5}
\end{equation}
Where $\otimes$ denotes the element-wise product.
\subsection{Enhance space relation loss for fine-grained image retrieval}
Asymmetric deep hash learning based image retrieval has achieved the promising retrieval accuracy and efficient training outperform other symmetric deep hash method in large-scale database. The query images are directly encoded into hash code via deep learning network. On the other hand, the database hash code is optimized bit by bit based on the fixed query hash code which is generated by the previous structure. The essence of this asymmetric deep hashing strategy is that finds out the relation between the query hash code latent space and the database hash code latent space so that it can preserve the similarity between the query hash code and database hash code as much as possible. However, when it comes to the fine-grained image retrieval task, the relation between the two latent spaces is vulnerable and implicit because the high intra-class variance and the low inter-class variance have a great impact on the low dimension hash code.

\begin{figure}
    \centering
    \includegraphics[scale=0.6]{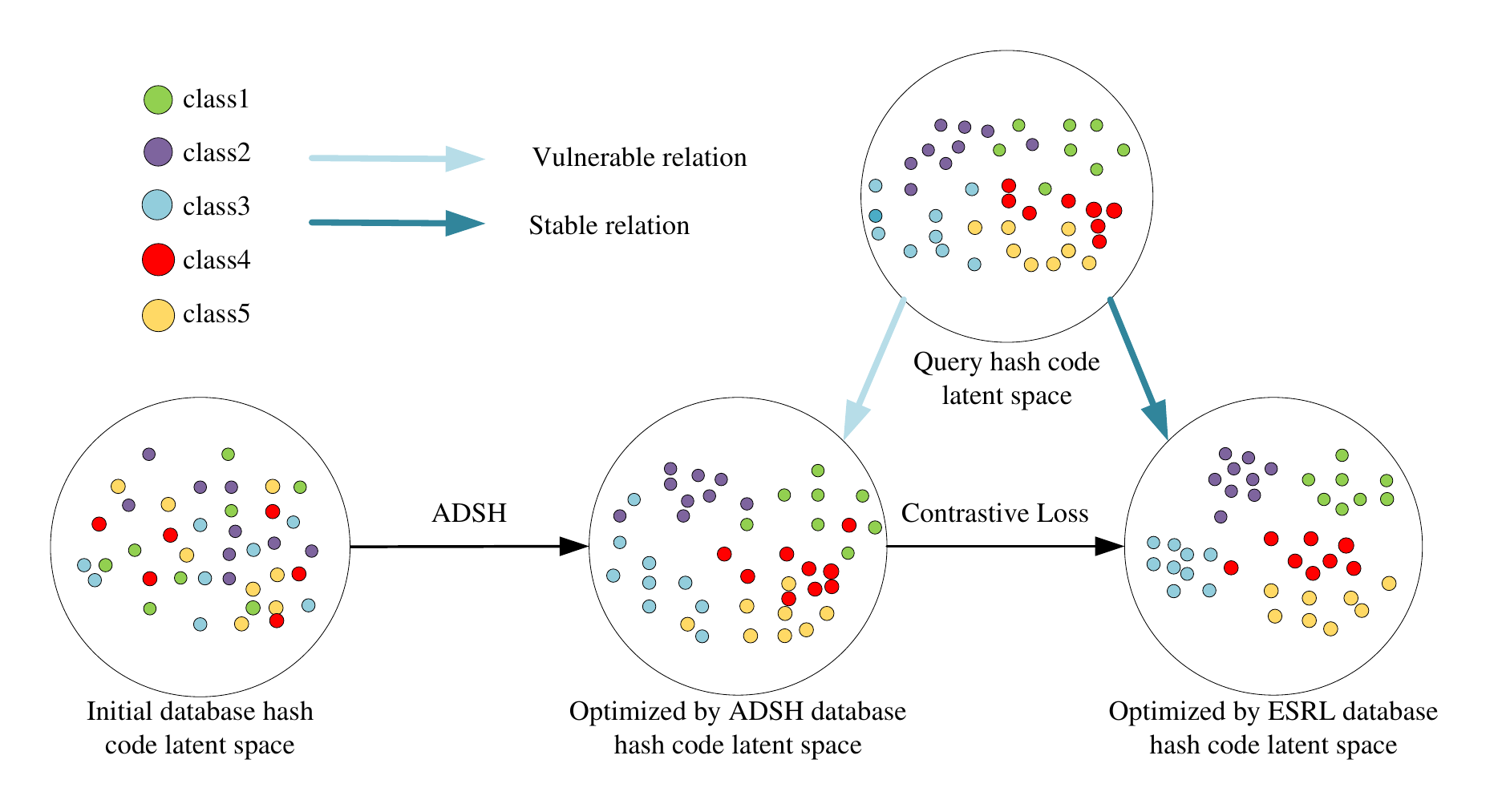}
    \caption{The illustration of the ESRL effect. By adding contrastive loss among the hash code triplets, we make the relation between query hash code latent space and database hash code latent space stabler to preserve the pair-wise similarity better.}
    \label{Fig.6}
\end{figure}

\par To address above problem, we introduce ESRL based on the contrastive loss about the triplet $ u_{tri}=\left\{{{u_i},{u_i{}^p},{\tilde{u_i}}}\right\}$. Based on the pair-wise similarity information and adaptive data augmentation with raw image, we make the category cluster of database hash code more compact by improving the vulnerable relation. After obtaining a stabler relation, the similarity between query hash code and database code is preserved better. The aim of enhancing space relation is shown in Fig.~\ref{Fig.6}. \cite{hadsell2006dimensionality} proposed the contrastive loss in Eq.~\ref{eq6} to enlarge the dissimilarity of inter-class and the similarity of intra-class.
\begin{equation}
    L(W) = \sum_{i=1}^P {\frac{1}{2}(1-Y)(D_W)^2 + \frac{1}{2}Y(max\left\{0, m-D_W\right\})^2}
    \label{eq6}
\end{equation}
Where $ {D_W}$ denotes the euclidean distance and ${m > 0}$ is a margin. $ Y$ denotes that the input pair is positive or not. 
\par However, the fine-grained information only includes the pair-wise similarity information rather than exact image label annotations so that enhancing loss function is employed as the metric of the positive pairs ${ {\left \{{u_i},{u_i{}^p} \right\}}}$, $ {\left \{u_i, {\tilde{u_i}} \right\}}$ from the $ u_{tri}$. ESRL $ L_{E}$ is in Eq.~\ref{eq9}.
\begin{equation}
    L_{self}(U, \tilde{U}) = \sum_{i=1}^r {\sum_{j=1}^k {(u_i{}_j - \tilde{u_i{}_j})^2}}
    \label{eq7}
\end{equation}
\begin{equation}
    L_{others}(U, U^P) = \sum_{i=1}^r {\sum_{j=1}^k {(u_i{}_j - {u_i{}_j{}^p})^2}}
    \label{eq8}
\end{equation}

\begin{equation}
    L_E(U,\tilde{U},U^P) = \alpha \cdot L_{self}(U, \tilde{U}) + \beta \cdot L_{others}(U, U^P)
    \label{eq9}
\end{equation}
Where $ r$, $ k$ denote the number of sampling images and the length of the hash code. $ {\alpha>0, \beta>0}$ are two hyper-parameters to control the different contributions to the final loss $ L$ for different hash code length. 

\subsection{Asymmetrical iteration optimizing}
Based on the previous architecture of FCAENet, we extract the hash code $ u_{tri}=\left\{{{u_i},{u_i{}^p},{\tilde{u_i}}}\right\}_{}$ based on $ r$ triplets. For the motivation of preserving similarity of query and database hash code, we adopt the Eq.~\ref{eq11} loss function:

\begin{equation}
    L_{sq}(U, V) = \sum_{i=1}^r {\sum_{j=1}^n {(u_i{}^{\top}{v_j}} - k \times {s_i{}_j})^2}
    \label{eq10}
\end{equation}

\begin{equation}
    L(U,V,\tilde{U},U^P) =  L_{sq}(U,V) + L_E(U, \tilde{U}, U^P) 
    \label{eq11}
\end{equation}
Where $ s_{ij}$ is the pair-wise similarity information. $ k$ is the hash code length. Combining the Eq.~\ref{eq7} to Eq.~\ref{eq11}, we can transform Eq.~\ref{eq11} to the Eq.~\ref{eq12}. From the Eq.~\ref{eq12}, the goal of minimizing the $ L$ is equal to making the three distance parts as close to zero as possible due to $ {\alpha>0, \beta>0}$.

\begin{equation}
\begin{aligned}
    L(u_i, \tilde{u_i}, u_i{}^p, v_j) = \sum_{i=1}^r {\sum_{j=1}^n {(u_i{}^{\top}{v_j}} - k \times {s_i{}_j})^2} + \\
    \alpha \sum_{i=1}^r {(u_i-\tilde{u_i})^2} + \beta \sum_{i=1}^r {(u_i-u_i{}^p)^2 }
    \label{eq12} 
\end{aligned}
\end{equation}

\par Set the sample images as $ X = \left \{X\right\}_{i=1}^r$ so that we denote $\Omega =\left \{X, \tilde{X}, X^P \right\}$.  We obtain the other form of the loss function in Eq.~\ref{eq13}, based on the  Eq.~\ref{eq1}, Eq.~\ref{eq2},  Eq.~\ref{eq7},  Eq.~\ref{eq8} and Eq.~\ref{eq12}.

\begin{equation}
\begin{aligned}
     \min_{\Theta, V} L(\Omega) = L_{sq}(U,V;S) + \alpha L_{self}(U, \tilde{U}) + \beta L_{other}(U, U^P) \\
    \mathrm{s.t.} \forall x_{tri}=\left\{x_i, \tilde{x_i},x^p\right\}, z_{tri} = f(x_{tri}),\\ u_{tri}=H(z_{tri},\theta), V \in \left\{-1,+1 \right\}^{n \times k}
\end{aligned}
\label{eq13}
\end{equation}
The activation function of the hash layer aims to generate hash code. Sign function is a great solution for hash layer but it is non-differentiable. Hence, we replace ${sign(\cdot)}$ with $ { \tanh(\cdot)} $ as the activation function.

\par In order to obtain the optimal solution of $ L(\Omega)$, we employ the two-step algorithm to learn $ {V, \Theta}$. The first step we fix $ V$ and learn $\Theta$. We exploit back-propagation algorithm to optimize the $\Theta$. After the $ \Omega$ finishing a forward process, we compute its gradient in Eq.~\ref{eq14}:

\begin{equation}
    \nabla_{\Theta}L(\Omega) = \nabla_{\Theta}L_{sq}(U,V;S) + \alpha \nabla_{\Theta} L_{self}(U, \tilde{U}) + \beta \nabla_{\Theta} L_{other}(U, U^P)
    \label{eq14}
\end{equation}
Then, we update the $\Theta$ in Eq.~\ref{eq15}:
\begin{equation}
    {\Theta}^{'}=\Theta - \eta \nabla_{\Theta}L(\Omega)
    \label{eq15}
\end{equation}
Where $\eta$ is the learning rate of the network.
\par The second step we fix $ \Theta$ and learn $ V$.When it comes to the $ \Theta$ fixed, $ {L_{self}(U, \tilde{U})}$ and $ {L_{other}(U, \tilde{U})}$ are viewed as a a constant so that the Eq.~\ref{eq13} can be write as Eq.~\ref{eq16}. 
\begin{equation}
\begin{aligned}
        L(V) &= L_{sq}(U,V;S) + {const}  \\
        &= \sum_{i=1}^r {\sum_{j=1}^n {(u_i{}^{\top}{v_j}} - k \times {s_i{}_j})^2} + {const} \\
        &= {{\Vert UV^{\top} \Vert}_F}^2-2ktr(S^{\top}UV^{\top}) + const, \mathrm{s.t.} V \in \left\{-1,+1 \right\}^{n \times k}
\end{aligned}
    \label{eq16}
\end{equation}
So, we can get the optimal solution of Eq.\ref{eq16} via ADSH\cite{jiang2018asymmetric} in Eq.\ref{eq17}.
    
\begin{equation}
    V_{*m} = -sign(\hat{V}_{m} \hat{U}^{\top}_{m}U_{*m}-2k{Q}_{*m})
    \label{eq17}
\end{equation}
We note the $ {{U}_{*m},{V}_{*m},{Q}_{*m}}$ denote the $ m$-th column of $ {U,V,Q}$ and $ {\hat{U}_m,\hat{V}_m,\hat{Q}_m}$ denote the matrix of $ {U,V,Q}$ excluding $ {{U}_{*m},{V}_{*m},{Q}_{*m}}$.

\section{Experiments}
\label{section4}

\subsection{Datasets}
We conduct experiments on two typical fine-grained datasets: CUB2011, Aircraft and three large-scale fine-grained datasets: NABirds, VegFru, Food101 to compare with other fine-grained image retrieval method.
CUB2011 contains 200 bird categories with 11788 images whose train part has 5994 images and test part has 5794 images. Aircraft contains 100 aircraft categories and total 10000 images that is split into 6667 images for training and 3333 images for testing. NABirds is a larger fine-grained bird dataset than CUB2011. It contains 555 bird categories and total 48562 images which is split into 23929 images for training and 24633 images for testing. VegFru is another large-scale fine-grained benchmark dataset for vegetables and fruits.There are 292 categories and total 146131 images with 29200 images(100 images per category) for training and 116931 for testing.The last large-scale fine-grained dataset Food101 contains 101 food categories and total 101000 images which 75750 images consist the training part and 25250 images are used for testing part. 

\subsection{Implementation details}
For a fair comparison of FCAENet with the previous works, our experiment settings are strict. In this paper, we use ResNet50 as baseline feature extractor and full-connection layer with non-linear activation function as hash layer. During experiments, we employ the same settings for baseline and FCAENet. We choose stochastic gradient descent (SGD) with momentum of 0.9 and weight decay of 0.00001 as optimizer. The initial learning rate is set to 0.001 and the mini-batch size is set to 64. As for 12bits and 24bits hash code, we train them for 40 iterations and the learning rate will be divided by 10 at iteration 20 and 30. As for 32bits and 48bits hash code, we train them for 60 iterations and the learning rate will be divided by 10 at epoch 30, 45 and 55. Each iteration includes 20 training epochs and each epoch handles $r$ images or triplets  that are randomly sampled from training dataset.
We set the same r as the ExchNet, 2000 samples for CUB2011, Aircraft, Food101 and 4000 samples for NABirds and VegFru due to the difficulty of the specific dataset. We resize the input image to $224 \times 224$ for training and testing. Finally, we just input one query image into FCAENet for inference rather than a triplet for training process so that we increase the retrieval speed with a pure but robust network.

% \begin{table}[]
%     \centering
%     \begin{tabular}{c|c}
%          &  \\
%          & 
%     \end{tabular}
%     \caption{Caption}
%     \label{tab1}
% \end{table}

% Please add the following required packages to your document preamble:
% \usepackage[table,xcdraw]{xcolor}
% If you use beamer only pass "xcolor=table" option, i.e. \documentclass[xcolor=table]{beamer}
% Please add the following required packages to your document preamble:
% \usepackage{multirow}
% Please add the following required packages to your document preamble:
% \usepackage{multirow}

% Please add the following required packages to your document preamble:
% \usepackage{multirow}

\subsection{Experimental result}
We adopt the retrieval accuracy as the evaluation metrics for a fair comparison with previous method. We calculate the retrieval accuracy by Eq.~\ref{eq18}.

\begin{equation}
     MAP = \frac{1}{N} \sum_{i=1}^N {AP}, AP = \frac{1}{n}\sum_{j=1}^K {\frac{n_j}{j}\times {pos(j)}}
    \label{eq18}
\end{equation}
Where $N$ denotes the number of the query images. $n$ denotes all of the correct retrieval images number in $K$. $n_j$ denotes all of the correct retrieval images number in $j$. And the $pos(j)$ denotes that if the $j$-th image is same as the query image, $pos(j)$=1, otherwise $pos(j)$=0.  
\par
We conduct extensive experiments based on four hash code length(12bits, 24bits, 32bits, 48bits) for five fine-grained datasets to show the performance of FCAENet. Our experimental results are illustrated in Table~\ref{tab1}. 

%\subsubsection{Fine-grained hashing on CUB-200-2011 dataset}
\par
On CUB2011 dataset, we achieve the 9.62\%, 8.69\%, 6.11\%, 9.09\% better than existing best retrieval performance of ExchNet \cite{cui2020exchnet} corresponding to different hash code length. On Aircraft dataset, we outperforms the ExchNet \cite{cui2020exchnet} 10.65\%, 29.63\%, 29.78\%, 22.29\% corresponding to different hash code. Besides, on other three large-scale fine-grained datasets, we still achieve the promising results comparing with ExchNet \cite{cui2020exchnet} and our baseline. On NABirds dataset, we get 7.34\%, 8.21\%, 9.64\%, 14.93\% MAP improvements than ExchNet \cite{cui2020exchnet} on 12bits, 24bits, 32bits, 48bits respectively. On VegFru dataset, we obtain 14.43\%, 19.19\%, 10.47\% MAP improvements than ExchNet \cite{cui2020exchnet} on 24bits, 32bits, 48bits respectively. Finally, on Food101 dataset, we achieve 21.08\%, 14.98\%, 18.95\% MAP improvements than ExchNet \cite{cui2020exchnet} on 24bits, 32bits, 48bits and comparable on 12bits.
% Please add the following required packages to your document preamble:
% \usepackage{multirow}

\begin{table}[]
\setlength{\abovecaptionskip}{0.cm}
\setlength{\belowcaptionskip}{0.3cm}
\centering
\caption{Retrieval accuracy(MAP) on fine-grained datasets.}
\resizebox{\textwidth}{30mm}{
\begin{tabular}{c|l|l|l|l|l|l|l|l|l|l|l|l|l|l}
\hline
\multicolumn{1}{l|}{Method} & \#Bits & LSH\cite{cui2020exchnet}    & SH\cite{cui2020exchnet}     & ITQ\cite{cui2020exchnet}     & SDH\cite{cui2020exchnet}     & DPSH\cite{cui2020exchnet}    & DSH\cite{cui2020exchnet}     & HashNet\cite{cui2020exchnet} & FPH\cite{yang2019feature}     & DaSH\cite{jin2020deep}   & ADSH\cite{cui2020exchnet}    & ExchNet\cite{cui2020exchnet} & \textbf{Baseline} & \textbf{FCAENet}    \\ \hline
\multirow{4}{*}{CUB2011}        & 12bits & 2.26\% & 5.55\% & 6.80\%  & 10.52\% & 8.68\%  & 4.48\%  & 12.03\% & -       & 14.2\% & 20.03\% & 25.14\% & {31.19\%}    & \textbf{34.76\%} \\
                            & 24bits & 3.59\% & 6.72\% & 9.42\%  & 16.95\% & 12.51\% & 7.97\%  & 17.77\% & -       & 20.9\% & 50.33\% & 58.98\% & {63.26\%}    & \textbf{67.67\%} \\
                            & 32bits & 5.01\% & 7.63\% & 11.19\% & 20.43\% & 12.74\% & 7.72\%  & 19.93\% & 58.32\% & -      & 61.68\% & 67.74\% & \textbf{74.13\%}    & {73.85\%} \\
                            & 48bits & 6.16\% & 8.32\% & 12.45\% & 22.23\% & 15.58\% & 11.81\% & 22.13\% & 61.24\% & 28.5\% & 65.43\% & 71.05\% & \textbf{80.77\%}    & {80.14\%} \\ \hline
\multirow{4}{*}{Aircraft}   & 12bits & 1.69\% & 3.28\% & 4.38\%  & 4.89\%  & 8.74\%  & 8.14\%  & 14.91\% & -       & -      & 15.54\% & 33.27\% & {39.48\%}    & \textbf{43.92\%} \\
                            & 24bits & 2.19\% & 3.85\% & 5.28\%  & 6.36\%  & 10.87\% & 10.66\% & 17.75\% & -       & -      & 23.09\% & 45.83\% & \textbf{76.78\%}    & {75.46\%} \\
                            & 32bits & 2.38\% & 4.04\% & 5.82\%  & 6.90\%  & 13.54\% & 12.21\% & 19.42\% & -       & -      & 30.37\% & 51.83\% & {81.13\%}    & \textbf{81.61\%} \\
                            & 48bits & 2.82\% & 4.28\% & 6.05\%  & 7.65\%  & 13.94\% & 14.45\% & 20.32\% & -       & -      & 50.65\% & 59.05\% & \textbf{82.21\%}    & {81.34\%} \\ \hline
\multirow{4}{*}{NABirds}    & 12bits & 0.90\% & 2.12\% & 2.53\%  & 3.10\%  & 2.17\%  & 1.56\%  & 2.34\%  & -       & -      & 2.53\%  & 5.22\%  & 11.88\%                   & \textbf{12.56\%}       \\
                            & 24bits & 1.68\% & 3.14\% & 4.22\%  & 6.72\%  & 4.08\%  & 2.33\%  & 3.29\%  & -       & -      & 8.23\%  & 15.69\% & 20.45\%                   & \textbf{23.90\%}       \\
                            & 32bits & 2.43\% & 3.71\% & 5.38\%  & 8.86\%  & 3.61\%  & 2.44\%  & 4.52\%  & -       & -      & 14.71\% & 21.94\% & 28.86\%    & \textbf{31.58\%}       \\
                            & 48bits & 3.09\% & 4.05\% & 6.10\%  & 10.38\% & 3.20\%  & 3.42\%  & 4.97\%  & -       & -      & 25.34\% & 34.81\% & 47.13\%                   & \textbf{49.74\%}       \\ \hline
\multirow{4}{*}{VegFru}     & 12bits & 1.28\% & 2.36\% & 3.05\%  & 5.92\%  & 6.33\%  & 4.60\%  & 3.70\%  & -       & -      & 8.24\%  & \textbf{23.55}\% & {17.58\%}    & {21.76\%} \\
                            & 24bits & 2.21\% & 4.04\% & 5.51\%  & 11.55\% & 9.05\%  & 8.91\%  & 6.24\%  & -       & -      & 24.90\% & 35.93\% & {45.40\%}    & \textbf{50.36\%} \\
                            & 32bits & 3.39\% & 5.65\% & 7.48\%  & 14.55\% & 10.28\% & 11.23\% & 7.83\%  & -       & -      & 36.53\% & 48.27\% & {65.62\%}    & \textbf{67.46\%} \\
                            & 48bits & 4.51\% & 6.56\% & 8.74\%  & 16.45\% & 9.11\%  & 17.12\% & 10.29\% & -       & -      & 55.15\% & 69.30\% & {79.39\%}    & \textbf{79.76\%} \\ \hline
\multirow{4}{*}{Food101}    & 12bits & 1.57\% & 4.51\% & 6.46\%  & 10.21\% & 11.82\% & 6.51\%  & 24.42\% & -       & -      & 35.64\% & \textbf{45.63}\% &  {39.31\%}    & {44.97\%} \\
                            & 24bits & 2.48\% & 5.79\% & 8.20\%  & 11.44\% & 13.05\% & 8.97\%  & 34.48\% & -       & -      & 40.93\% & 55.48\% & {73.88\%}    & \textbf{76.56\%} \\
                            & 32bits & 2.64\% & 5.91\% & 9.70\%  & 13.36\% & 16.41\% & 13.10\% & 35.90\% & -       & -      & 42.89\% & 56.39\% & {79.88\%}    & \textbf{81.37\%} \\
                            & 48bits & 3.07\% & 6.63\% & 10.07\% & 15.55\% & 20.06\% & 17.18\% & 39.65\% & -       & -      & 48.81\% & 64.19\% & {81.98\%}    & \textbf{83.14\%} \\ \hline
\end{tabular}}
\label{tab1}
\end{table}

\par During the experiments, we find that the results of our baseline model improve than the notable method, ExchNet \cite{cui2020exchnet}. About this situation, we make carefully analysis in the Sec.~\ref{section4.5.2}.

\subsection{Ablation study}
In the ablation study, we conduct three components ablation study. The first component is that we adopt SREM and ESRL to improve the performance of fine-grained hashing. The second component part, we conduct on the selection of different region numbers to erase. The last component, we compare the different region size influencing on retrieval accuracy. 

% Please add the following required packages to your document preamble:
% \usepackage[table,xcdraw]{xcolor}
% If you use beamer only pass "xcolor=table" option, i.e. \documentclass[xcolor=table]{beamer}
\begin{table}[]
\setlength{\abovecaptionskip}{0.cm}
\setlength{\belowcaptionskip}{0.3cm}
\centering
\caption{Ablation comparison experiments of SREM and ESRL on Food101 dataset.}
\begin{tabular}{l|l|l|l|l}
\hline
Bits\#    & 12bits & 24bits & 32bits & 48bits \\ \hline
Baseline  & 39.31\% & 73.88\% & 79.88\% & 81.98\% \\
Baseline + SREM  & 43.58\%       & 75.67\%       & 80.25\%       & 81.99\%       \\ 
Baseline + ESRL & 43.67\% & 75.24\%       & 81.11\%       & 82.41\% \\ \hline
FCAENet & \textbf{44.97}\% & \textbf{76.56}\% & \textbf{81.37}\% & \textbf{83.14}\% \\ \hline
\end{tabular}
\label{tab2}
\end{table}

\subsubsection{Ablation study of SREM and ESRL}
To alleviate the impact of high intra-class variance and enhance the relation between query hash latent space and database hash latent space, we introduce the SREM and ESRL respectively. To verify that SREM and ESRL are really contributed to promote the performance of fine-grained hashing, we conduct the experiments for w/ or w/o the SREM and ESRL on the Food101 dataset with four hash code length. Table~\ref{tab2} shows the image retrieval performance of w/ or w/o the SREM and ESRL.
\par Compared to our baseline, employing only SREM improves  4.27\%, 1.79\%, 0.37\%, 0.01\% retrieval accuracy on 12bits, 24bits, 32bits, 48bits hash code. When we use the ESRL, the retrieval accuracy is promoted 4.36\%, 1.36\%, 1.23\%, 0.43\% on 12bits, 24bits, 32bits, 48bits hash code. From the results of Table~\ref{tab2}, the improvements on 32bits or 48 bits is not obvious  by contrast to the 12bits or 24bits. This is because the representative ability of short bit hash code is poor, so that there is some prominent discrimination between the optimized network and baseline network. However, as for 32bits or 48bits, their abilities of representation are decent. Hence, the improvement of retrieval accuracy is not dramatic so much.

\subsubsection{Ablation study of the region number of SREM}
The influence of different number of erasing region is shown in Table~\ref{tab3}. In this experiment, we choose four kinds of region numbers.
It can be seen in Table~\ref{tab3} that, compared with different region numbers, choosing proper region number has influence on the performance of SREM. By contrast to the four different region numbers, 50 shows the best performance among the region numbers. Fine-grained task needs to pay attention to some subtle regions. Hence, if the number of erasing region is too few, the SREM hardly works for promoting the image retrieval accuracy. On the contrary, large number of erasing regions drop out almost significant part so that they will make the retrieval performance reduce a lot. Hence, we select the 50 erasing regions for our experiments. 

\begin{table}[]
\setlength{\abovecaptionskip}{0.cm}
\setlength{\belowcaptionskip}{0.3cm}
\centering
\caption{Ablation comparison experiments of the region number of SREM on Food101 dataset based on 24bits hash code.}
\begin{tabular}{lllll}
\hline
\multicolumn{1}{l|}{Number} & 10 & 50 & 100 & 150 \\ \hline
\multicolumn{1}{l|}{FCAENet}   & 75.23\%     & \textbf{76.56\%}   & 76.41\%   & 74.84\%   \\ \hline
                            &    &    &    &   
\end{tabular}
\label{tab3}
\end{table}

\begin{table}[]
\setlength{\abovecaptionskip}{0.cm}
\setlength{\belowcaptionskip}{0.3cm}
\centering
\caption{Ablation comparison experiments of the region size of SREM on Food101 dataset based on 24bits hash code.}
\begin{tabular}{lllll}
\hline
\multicolumn{1}{l|}{Size} & 16 & 32 & 64 & 128 \\ \hline
\multicolumn{1}{l|}{FCAENet}   & 75.66\%    & \textbf{76.56\%}   & 76.30\%    & 74.60\%   \\ \hline
                            &    &    &    &   
\end{tabular}
\label{tab4}
\end{table}

\subsubsection{Ablation study of the region size of SREM}
Different region size of SREM brings different retrieval accuracy. We also conduct experiments based on four region size of SREM, as is shown in Table~\ref{tab4}. We find that the image retrieval accuracy of 128 erasing region size decreases 1.94\% than 32. But the size of 16 brings 0.9\% accuracy reduction than 32. That because the large erasing region size covers the main part of the object and the small erasing region size makes the SREM no sense for complete architecture.

% Please add the following required packages to your document preamble:
% \usepackage{multirow}

% \begin{table}[]
% \centering
% \caption{Comparison unsupervised method like-SimCLR for Fine-grained hashing and our method on CUB dataset}
% \begin{tabular}{l|ll}
% \hline
% Dataset & \multicolumn{2}{c}{CUB} \\ \hline
% Bits\#  & 12bits      & 24bits     \\ \hline
% base1   & 0.0237      & 0.0204     \\
% ourbase & 0.3476      & 0.6767     \\ \hline
% \end{tabular}
% \label{tab7}
% \end{table}

\subsection{Analysis and visualization}

\begin{figure}[]
    \centering
    \includegraphics[width=12cm, height=8cm]{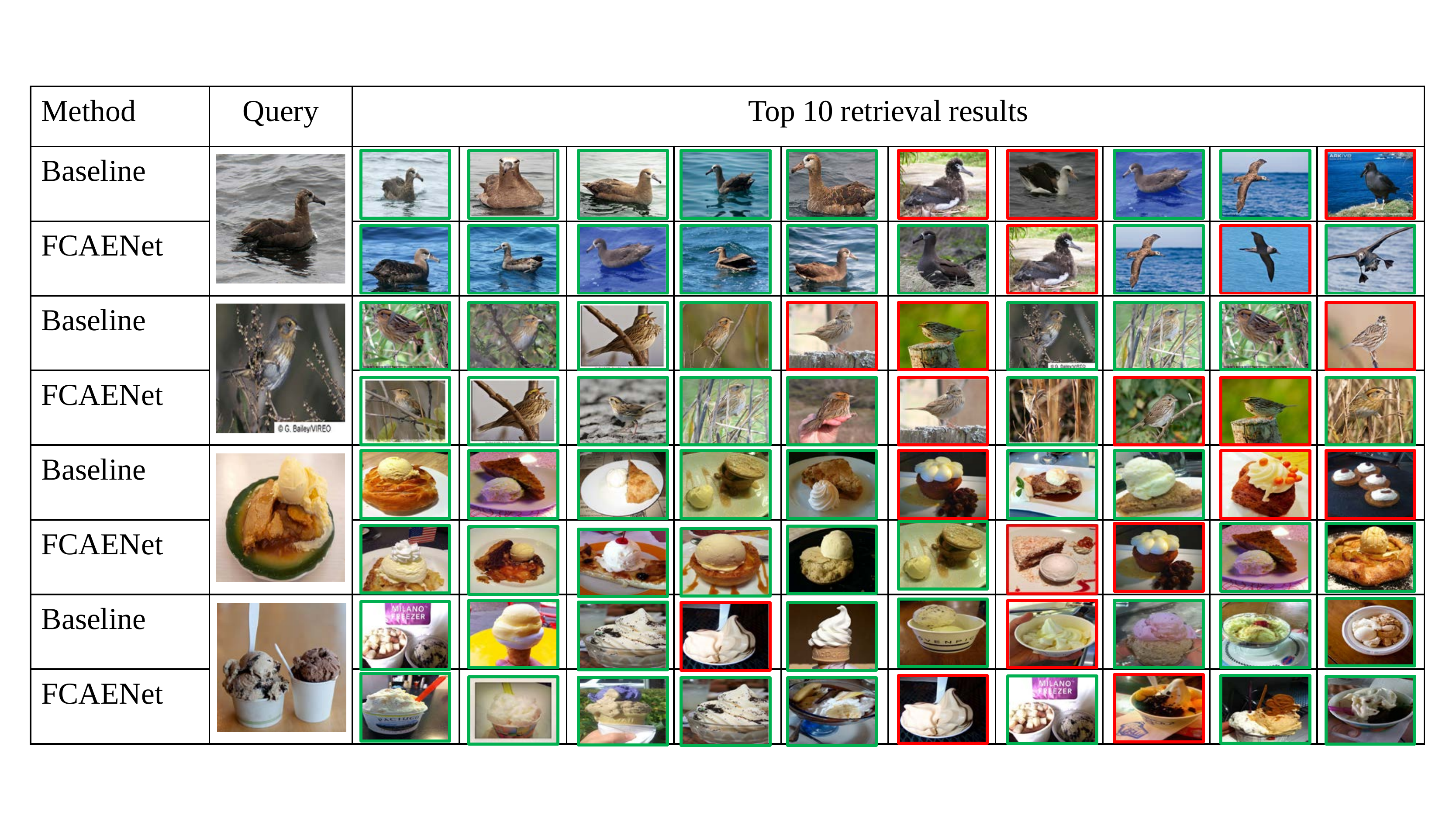}
    \caption{The top 10 retrieval results of FCAENet on fine-grained datasets.The green rectangle means correct retrieval result and the red rectangle means wrong retrieval result.}
    \label{Fig.6}
\end{figure}
\subsubsection{Retrieval result visualization}
To more intuitively show the performance of FCAENet, we visualize the top10 retrieval images in Fig.~\ref{Fig.6} on CUB2011 and Food101 datasets. Meanwhile, we compare the retrieval results between our baseline and FCAENet. As is shown in Fig.~\ref{Fig.6}, the green box denotes the correct retrieval result and the red box denotes the wrong retrieval result. As we can see for the first query images in Fig.~\ref{Fig.6}, the wrong retrieval results are hard to discriminate especially for the first row and the sixth column of retrieval result of our baseline. Its body shape and beak are almost same as the query images. However, after introducing FCAENet, we can find that the first wrong position becomes the seventh column of retrieval result. And the wrong image retrieval number reduces obviously. On the other hand, among the first query image retrieval results which are based on FCAENet, the eighth column and the tenth column show a significant difference with the query images due to the pose of the birds, but they exactly belong to the same category as the query image. Hence, FCAENet truly improves the retrieval accuracy on pulling the similar images closer and pushing the dissimilar images further via visualizing the top10 retrieval results.

\subsubsection{Our promising baseline analysis}
\label{section4.5.2}
\begin{figure}[]
    \centering
    \includegraphics[width=12.5cm, height=8cm]{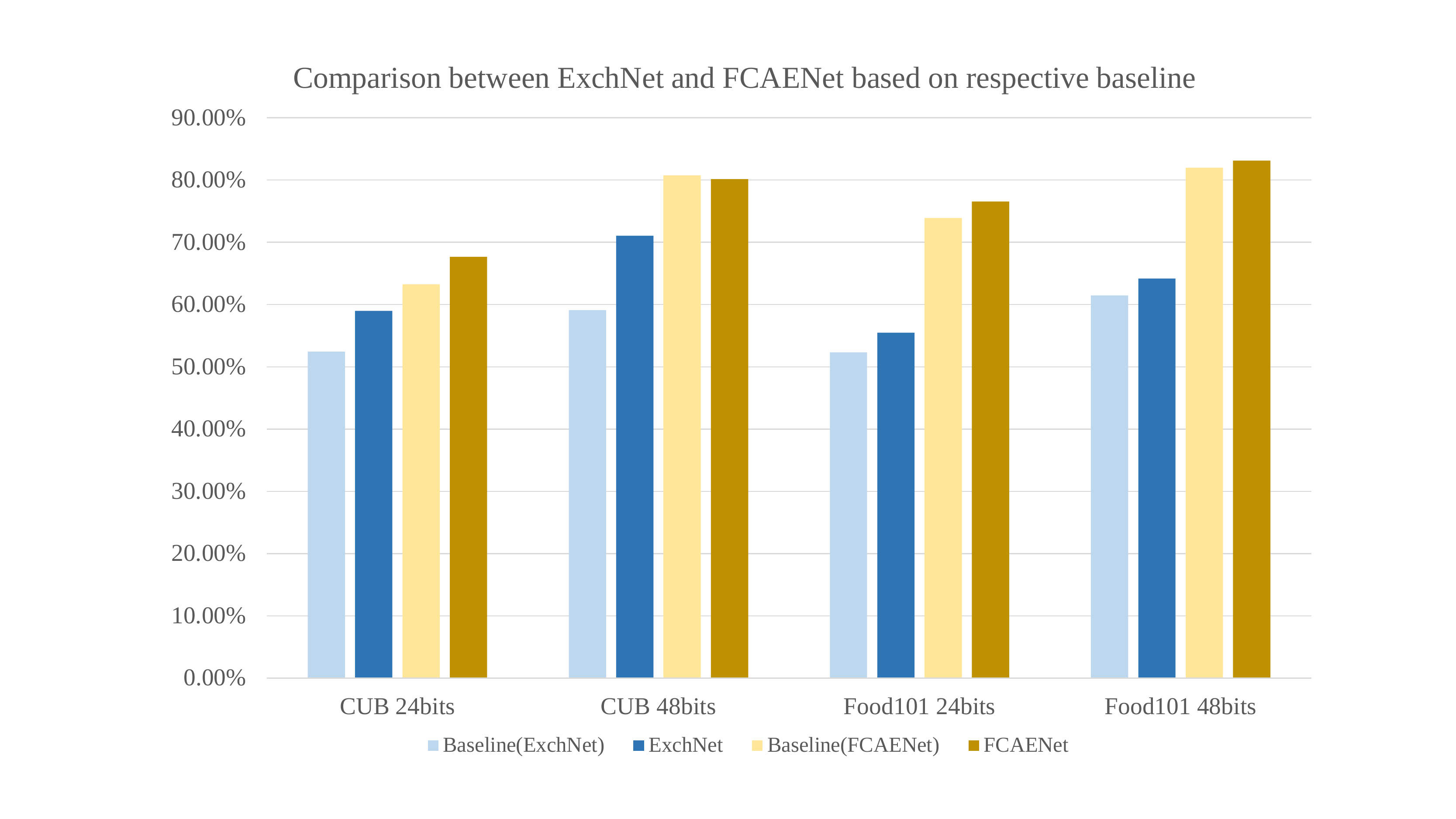}
    \caption{Comparison between ExchNet and FCAENet based on respective baseline.}
    \label{Fig.7}
\end{figure}

During the experiments, we achieve promising improvements on almost four hash code lengths than previous work. To verify our experiment results are reliable, we conduct the the experiment to find the reason on CUB2011 and Food101 datasets with 24bits hash code and 48bits hash code. The experiment results are shown in Table~\ref{tab5} and Fig.~\ref{Fig.7}. With comparison of the architecture of ExchNet\cite{cui2020exchnet}, we find out the main difference. We just employ the compete ImageNet pre-trained ResNet50 as backbone to extract feature vectors, but the ExchNet backbone employs the first three stages of ResNet50 and the fourth stage is used as a local attention generating module without downsampling. Based on this local attention module, ExchNet add the feature alignment module behind its backbone. There is no doubt that ExchNet brings an incredible improvement on fine-grained image retrieval than previous, but the local attention module which is in the backbone may carry a new challenge for representing with low dimensional hash code. For the above reasons, our baseline almost achieves great improvements than the previous method and FCAENet is proved exactly effective with comparison of our baseline. 

\begin{table}[]
\setlength{\abovecaptionskip}{0.cm}
\setlength{\belowcaptionskip}{0.3cm}
\centering
\caption{Comparison the retrieval accuracy between ExchNet and FCAENet based on respective baseline.}
\begin{tabular}{c|l|ll|ll}
\hline
\multicolumn{2}{c|}{Dataset}           & \multicolumn{2}{c|}{CUB2011} & \multicolumn{2}{c}{Food101} \\ \hline
\multicolumn{2}{c|}{Bits\#}            & 24bits      & 48bits     & 24bits        & 48bits       \\ \hline
\multicolumn{2}{c|}{ExchNet\_Baseline} & 52.43\%     & 64.65\%    & 49.46\%       & 57.23\%      \\ \cline{1-2}
\multicolumn{2}{c|}{ExchNet\cite{cui2020exchnet}}           & 58.98\%     & 71.05\%    & 55.48\%       & 64.19\%      \\ \cline{1-2}
\multicolumn{2}{c|}{Baseline(ours)}     & 63.26\%     & \textbf{80.77\%}    & 73.88\%       & 81.98\%      \\ \cline{1-2}
\multicolumn{2}{c|}{FCAENet}       & \textbf{67.67}\%     & 80.14\%    & \textbf{76.56\%}       & \textbf{83.14\%}      \\ \hline
\end{tabular}
\label{tab5}
\end{table}

\subsubsection{Visualization of SREM data augmentation}
\par
\begin{figure}[]
    \centering
    \includegraphics[width=12cm, height=6cm]{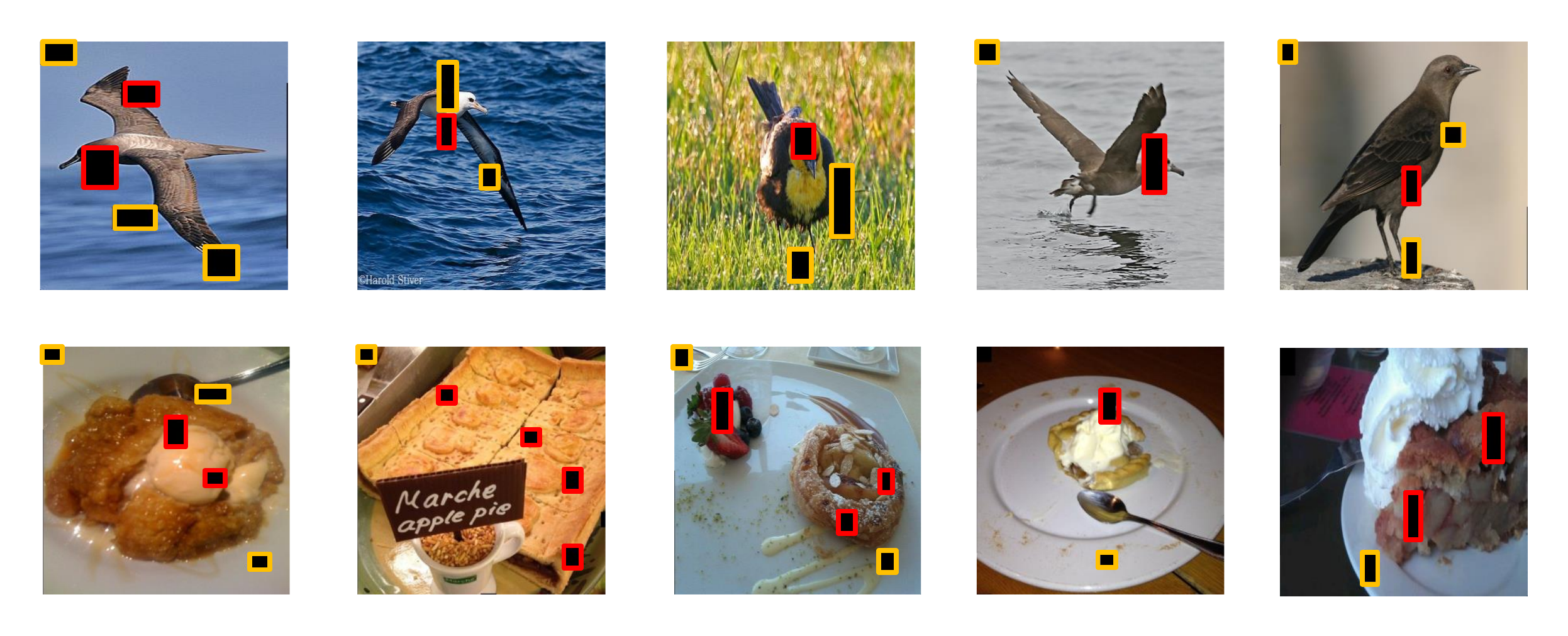}
    \caption{Visualization of erasing region of SREM. The red rectangles frame the discriminative local part. The yellow rectangles frame the global background information.}
    \label{Fig.8}
\end{figure}

The aim of design SREM is to make the FCAENet more robust for hashing based fine-grained image retrieval. From the outputs of SREM in Fig.~\ref{Fig.8}, we can intuitively find that FCAENet exactly finds the most discriminative parts. For example, the first image in Fig.~\ref{Fig.8} is one of a bird species and its significant parts may be the beak and wings which are exactly framed by the red rectangle. Besides, when the inter-class variance is too low to discriminate, the background information may make great contribution to correctly retrieval. The yellow rectangles in the above images show that FCAENet still pays attention to the global information. By integrating the local subtle features and the global background features, we exploit the attention region better than previous methods in fine-grained image retrieval problem. Therefore, the representation ability of low dimensional hash code become better than previous methods. By the way, we set the number of erasing region as 50 and the size of erasing region as 32 but we only view few erasing regions and different size region in Fig.~\ref{Fig.8}. That is because the positions of the hottest region are always consecutive or adjacent, so that many erasing regions intersect together. These visualization results illustrate that SREM exactly works for promoting the retrieval performance. 

\subsubsection{Visualization of feature map}
\begin{figure}[]
    \centering
    \includegraphics[width=12cm, height=5cm]{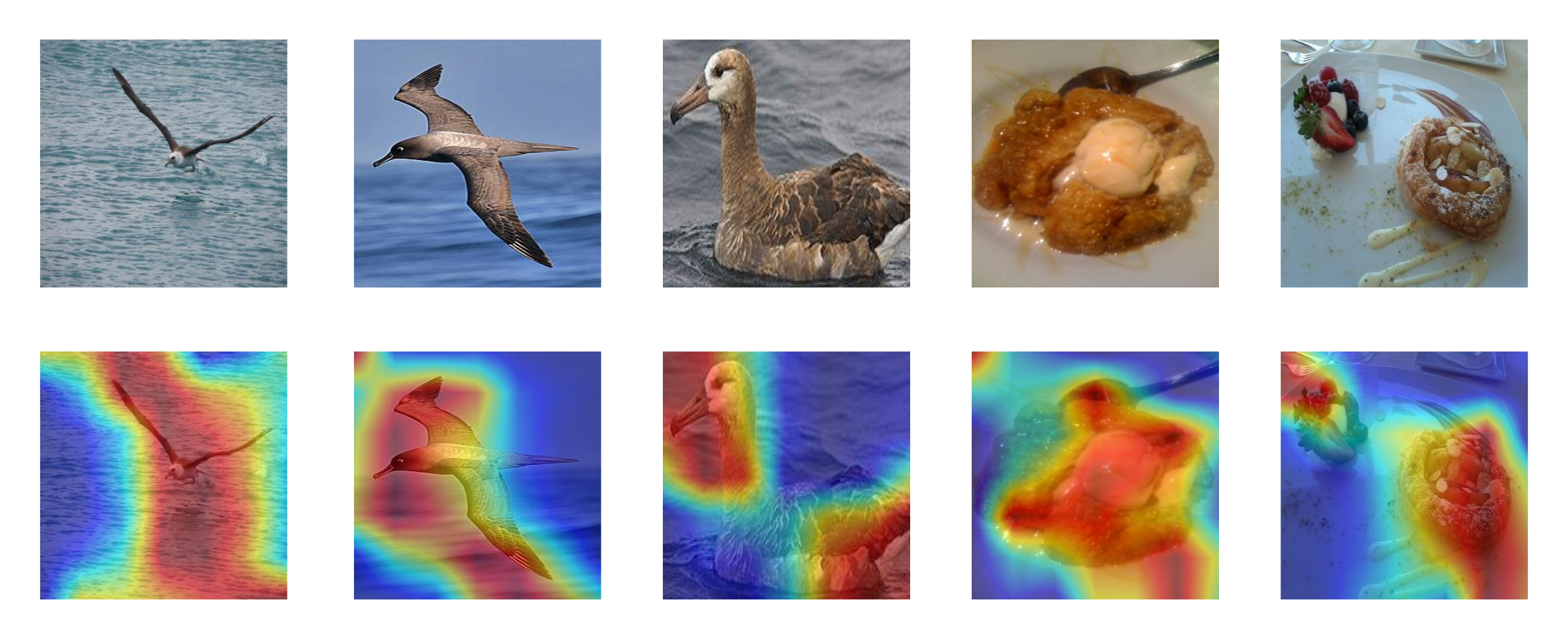}
    \caption{Visualization of attention map generated by the feature extractor of FCAENet}
    \label{Fig.9}
\end{figure}

To further verify that FCAENet can find a more convincing balance between the local discriminative information and global background information than previous method, we visualize the attention maps in Fig.~\ref{Fig.9}. From all of class activation maps in Fig.~\ref{Fig.9}, we can find that not only the local parts of object are focused on by the network but also some background regions are also preserved by the architecture. For example, there is a common character that although the image corner often contains less object information, it is always the highlight region of the feature map. These visualization results show that just use the local part information with much information loss in low dimensional hash code space. Hence, making full use of background information and local part information will result the performance of fine-grained image retrieval better.

\section{Conclusion}
\label{section5}
In this paper, we propose FCAENet to improve the performance of fine-grained image retrieval based on hash learning. The selective region erasing module(SREM) makes the feature extractor obtain more discriminative low dimensional hash code for fine-grained image retrieval. The enhance space relation loss(ESRL) makes the relations between query hash code latent space and database hash code latent space stabler, so that the same category images can be mapped into the same cluster in low dimensional hash latent space. Depending on the cooperation between the SREM and ESRL, FCAENet achieves new state-of-the-art retrieval accuracy on CUB2011, Aircraft, VegFru, Food101, NABirds datasets. Especially on the short hash code, the retrieval performance achieves promising improvements. As for the large-scale fine-grained image retrieval, attaining the pair-wise similarity information may cost a lot. Currently, the unsupervised representation learning has achieved great success. But these unsupervised researches based on a normal embeddings size rather than a certain hash code. Hence, future work may focus on combining the unsupervised representation learning and deep hashing learning for fine-grained image retrieval task. 

\section*{Acknowledgments}
This work was supported by National Natural Science Foundation of China under Grants 62002005 and 62072021.

\bibliography{FCAENet}

\end{document}